\documentclass[aos,preprint]{imsart}
\setattribute{journal}{name}{} 

\RequirePackage[OT1]{fontenc}
\RequirePackage{amsthm,amsmath,amssymb}

\RequirePackage{natbib}
\setcitestyle{round,authoryear,citesep={;},aysep={,},yysep={;}}

\renewcommand{\cite}[1]{\citep{#1}}

\RequirePackage[colorlinks,citecolor=blue,urlcolor=blue]{hyperref}

\usepackage{graphicx} 

\usepackage{macros}

\usepackage{caption}
\usepackage{subcaption}

\usepackage{verbatim}
\usepackage{gensymb}

\usepackage[top=1in, bottom=1in, left=1in, right=1in]{geometry}

\usepackage{macros}
\usepackage{savesym}
\savesymbol{captionbox}
\usepackage[font={small}]{caption}
\DeclareCaptionType{copyrightbox}
\usepackage{subcaption}
\restoresymbol{CAP}{captionbox}

\usepackage{amsmath,amssymb}
\usepackage{verbatim}
\usepackage{gensymb}
\usepackage{multirow}
\usepackage{hhline}
\usepackage{wrapfig}
\usepackage{todonotes}

\begin{document}

\begin{frontmatter}
\title{Scalable Bayesian Inference for \\Excitatory Point Process Networks}
\runtitle{Scalable Bayesian Inference for Excitatory Point Process Networks}

\begin{aug}
\author{\fnms{Scott W.} \snm{Linderman}\ead[label=e1]{swl@seas.harvard.edu}}
\and
\author{\fnms{Ryan P.} \snm{Adams}\ead[label=e2]{rpa@seas.harvard.edu}}
\affiliation{Harvard University}

\runauthor{S. W. Linderman and R. P. Adams}
\end{aug}

\begin{abstract} 
Networks capture our intuition about relationships in the world. 
They describe the friendships between Facebook users, interactions in financial markets, and synapses connecting neurons in the brain.
These networks are richly structured with cliques of friends, sectors of stocks, and a smorgasbord of cell types that govern how neurons connect.
Some networks, like social network friendships, can be directly observed, but in many cases we only have an indirect view of the network through the actions of its constituents and an understanding of how the network mediates that activity.  
In this work, we focus on the problem of latent network discovery in the case where the observable activity takes the form of a mutually-excitatory point process, also known as a Hawkes process.
We build on previous work that has taken a Bayesian approach to this problem, specifying prior distributions over the latent network structure and a likelihood of observed activity given this network.
We extend this work by proposing a discrete-time formulation and developing a computationally efficient stochastic variational inference (SVI) algorithm that allows us to scale the approach to long sequences of observations.
We demonstrate our algorithm on the calcium imaging data used in the Chalearn neural connectomics challenge.
\end{abstract} 
\end{frontmatter}

\section{Introduction}

Networks are abstractions of the relationships and connections between real-world objects, such as people, stocks, or neurons.
These connections reflect relationships like ``Wilson and Brady are friends,'' or ``When neuron A fires it excites neuron B.'' 
Sometimes the networks themselves are observed data, as in the case of social network friendships, but often our view of the network is indirect.
We are left to infer the latent connections between objects based on our observations of their behavior.
In our neural example, recording techniques can often provide a measure of the neurons' activity but cannot resolve the individual synaptic connections between neurons.
Given our knowledge of how synapses work, however, we might infer that if one neuron consistently fires shortly after another then there is likely an excitatory connection between them.
This is one example of  the \emph{latent network discovery} problem that this work addresses.

We focus on the case where our observations come in the form of a series of discrete events, like a sequence of Twitter messages or the firing pattern of a population of neurons.
These events do not happen independently; rather, events induce subsequent events according to an excitatory network of interactions.
A connection from one object to another indicates that events by the first object increase the probability of subsequent events by the second.
We model these observations with a multivariate Hawkes process, a type of point process tailored to excitatory networks of interaction.

Building on the previous work, we combine the Hawkes process observation model with a prior distribution over networks in a unified Bayesian model \cite{Simma-2010, Blundell-2012, Perry-2013, Dubois-2013, Linderman-2014, Guo-2015}.
Most real-world networks are not simply random, but have highly structured patterns of interaction.
For example, a friendship can often be traced back to some commonality between two people, such as belonging to the same club or attending the same school.
A simple model for social networks might assign each person to a group, and then connect people according to whether or not they are in the same group.
This is known as a stochastic block model (SBM) \cite{Nowicki-2001}, and is one example of a random network model that may serve as prior probability distribution over networks in a Bayesian framework.

We improve upon previous work by providing a discrete-time analogue of the Hawkes process that is considerably more efficient on datasets with high rates of activity, and by devising an efficient stochastic variational inference (SVI) algorithm that can scale to long sequences of observations.
Most previous work has relied upon Markov chain Monte Carlo (MCMC) methods for inference,  which must consider the entire observation sequence when evaluating the likelihood of a state, and which can be prone to poor convergence.
SVI provides an alternative method of inference that can work with mini-batches (small subsets) of observations per iteration, and has been shown to yield dramatic improvements in a variety of large-scale machine learning problems. Code for the models and algorithms developed in this paper is available at \url{https://github.com/slinderman/pyhawkes}.

\begin{figure*}[t]
\begin{center}
\begin{subfigure}[b]{.32\textwidth}
\centering
\includegraphics[width=2.34in]{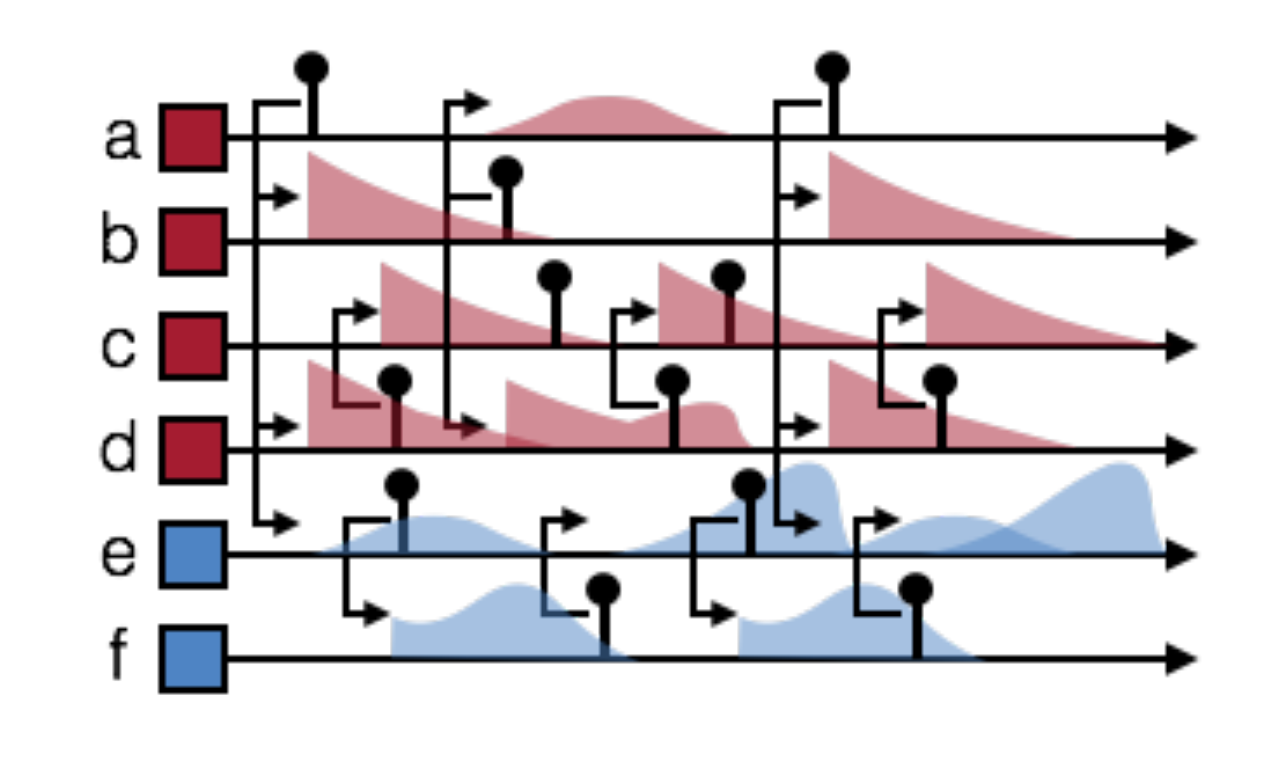} 
\caption{Hawkes process dynamics}
\label{fig:hawkes_a}
\end{subfigure}
\begin{subfigure}[b]{.32\textwidth}
\centering
\includegraphics[width=0.81in]{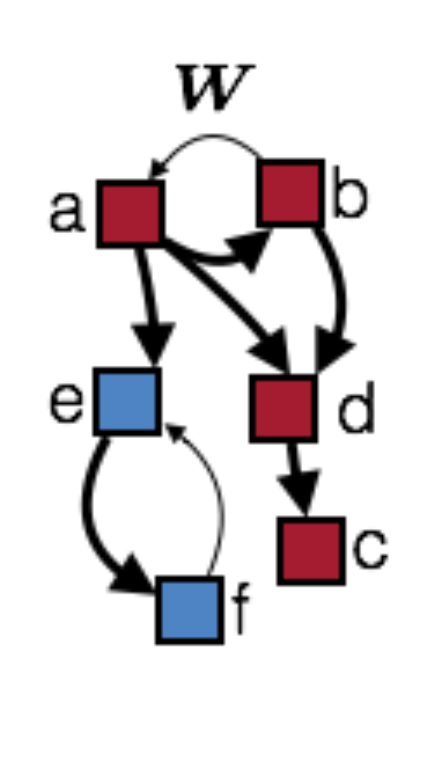} 
\caption{Weighted Network}
\label{fig:hawkes_b}
\end{subfigure}
\begin{subfigure}[b]{.32\textwidth}
\centering
\includegraphics[width=2.10in]{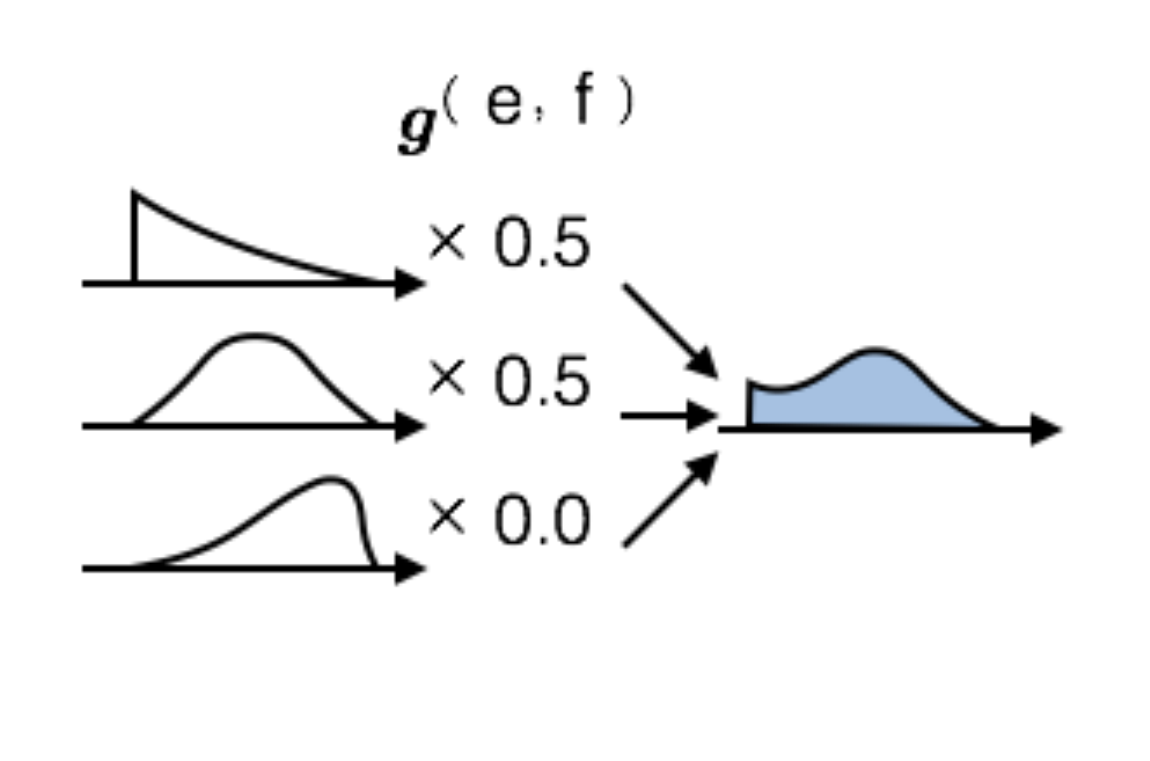} 
\caption{Impulse responses}
\label{fig:hawkes_c}
\end{subfigure}
\end{center}
\caption{Components of the Network Hawkes process model. (a) Events induce weighted impulse responses on downstream processes, spawning more events according to the Hawkes process dynamics. (b) Underlying these dynamics is a weighted, directed network. Here, the network is a stochastic block model with two clusters of processes (red and blue). Processes primarily connect to others of the same type, though there is a small probability of connection from red to blue. (c) We model the impulse responses as a convex combination of normalized basis functions.}
\label{fig:hawkes}
\end{figure*}

\section{Related Work}
Hawkes processes \cite{Hawkes-1971} are multivariate point processes that relate excitatory interaction networks to sets of discrete events. 
For example, suppose we have a Hawkes process with two constituent processes.
An event on the first process could add an \emph{impulse response} to the future rate of the second process.
More generally, a multivariate Hawkes process consists of~$K$ individual point processes with rates~${\{\lambda_k(t)\}_{k=1}^K}$ that depend upon the events that have occurred up to time~$t$.   
This dependence on preceding events distinguishes the Hawkes process from the Poisson process.
Given the history of events up to time~$t$, however, Hawkes processes have conditionally Poisson dynamics.

Hawkes processes have additive, excitatory interactions.
Each event adds a nonnegative impulse response to the future rate of connected processes.
The rate of the~$k$-th process is
\begin{align}
\label{eq:cont_hawkes_rate}
\lambda_{k}(t) &= \lambda_k^{(0)} + \sum_{k' = 1}^K \sum_{n=1}^{N_k} \bbI[s_{k,n}<t] \cdot h_{k' \to k}(t-s_{k,n}),
\end{align}
where~${\{s_{k,n}\}_{n=1}^{N_k} \in [0,T]^{N_k}}$ is the set of event times for events on process~$k$,~$N_k$ is the total number of events on process~$k$,~$\lambda_k^{(0)}$ is the ``background rate'' of the~$k$-th process, and~${h_{k' \to k}(\Delta t)}$ is an impulse response function describing the amplitude of influence that events on process~$k'$ have on the rate of process~$k$ at time lag~${\Delta t}$.

The Hawkes process is closely related to the generalized linear model (GLM) with Poisson observations, which is widely used in computational neuroscience \cite{Paninski-2004, Pillow-2008}.  
In fact, the Hawkes process is a special case of the Poisson GLM in which the link function is linear and the impulse responses are non-negative. 
As in the Poisson GLM, the negative log likelihood of the Hawkes process is convex, enabling efficient maximum likelihood and maximum \emph{a posteriori} estimation. 
The advantage of the Hawkes process is that its linear form allows for elegant fully-Bayesian inference --- a task which is non-trivial  in the Poisson GLM due to the lack of conjugacy. With these Bayesian inference algorithms, we can estimate and reason using the posterior uncertainty of the model.

One of the earliest applications of Hawkes processes in machine learning was the work of~\citet{Simma-2010}, which developed an expectation-maximization algorithm based upon the auxiliary variable parent formulation. Observing that~$\lambda_{k}(t)$ is a sum of impulse responses, \citet{Simma-2010} invoked the Poisson superposition theorem to motivate a set of auxiliary ``parent'' variables,~$z_{k,n}$, which denote the origin of the~$n$-th event on process~$k$, either the background rate or an impulse response from a previous event. Conditioned upon these auxiliary variables, the likelihood factorizes over impulse responses. 

Subsequent works leveraged this intuition to extend Hawkes processes with interpretable prior distributions over the network of impulse responses.
\citet{Blundell-2012} introduced an infinite relational model prior over the network of interactions as well as a Gibbs sampling algorithm for fully Bayesian inference.
\citet{Dubois-2013} explored the use of infinite relational models as a prior in conjunction with a point process observation model and a Gibbs sampling inference algorithm.
Recently,~\citet{Linderman-2014} developed a general framework for combining random ``spike-and-slab'' network models with Hawkes processes that uses a continuous time formulation and an auxiliary Gibbs sampling inference algorithm.
\citet{Guo-2015} have developed a similar model that applies Hawkes processes to language modeling problems and incorporates features of the discrete events.

We extend the work of \citet{Linderman-2014} by addressing two shortcomings: the limited scalability of the continuous time formulation which introduces auxiliary variables for each event in the dataset, and the batch nature of their Gibbs sampling algorithm. We address the former by deriving a discrete time version of their model which vastly outperforms the continuous time version on datasets with high rates of activity. To overcome the batch nature of the Gibbs algorithm, we make an approximation to the spike-and-slab network model that renders the model fully conjugate, thereby enabling efficient stochastic variational inference \cite{Hoffman-2013} on mini-batches of data.

\section{The Discrete Time Network Hawkes Model}
The fundamental limitation of the previously developed continuous time models is that the number of values that the auxiliary variable~$z_{k,n}$ can take grows with the number of events which occurred before time~$s_{k,n}$. For datasets with high rates of activity, this can quickly become the limiting factor of the inference algorithm.  At the same time, it is often reasonable to assume that events do not interact on time scales faster than some~$\Delta t$. This motivates a discrete time formulation in which we bin events in bins of width~$\Delta t$ and ignore potential interactions between events in the same bin. Then the rate becomes,
\begin{align}
\label{eq:discrete_hawkes_rate}
 \lambda_{t,k} &= \lambda_k^{(0)} + \sum_{k' = 1}^K \sum_{t'=1}^{t-1} s_{t',k'} \cdot h_{k' \to k}[t-t'],
\end{align}
where~${\bs \in \naturals^{T\times K}}$ is the matrix of event counts and~${h_{k' \to k}[t-t']}$ is an impulse response function describing the amplitude of influence that events on process~$k'$ have on the rate of process~$k$ at discrete time lag~${t-t'}$. As we will show, under this formulation the auxiliary variables only assume a fixed set of values independent of the rate.

In order to incorporate the network model as a prior distribution for the Hawkes process, we follow the approach of~\citet{Linderman-2014} and decompose the impulse response function into the product of a scalar weight that specifies the strength of the interaction and a probability mass function that specifies the time course of interaction:
\begin{align*}
h_{k \to k'}[d]=W_{k \to k'} G_{k \to k'}[d] \equiv W_{k \to k'} \sum_{b=1}^B g_{b}^{(k,k')} \phi_b[d],
\end{align*}
for~${d\in\{1,\ldots,D\}}$. 
Here,~${\bW\in\reals_+^{K\times K}}$ is a non-negative weight matrix drawn from a spike-and-slab prior,
\begin{align*}
A_{k \to k'} &\sim \distBernoulli(A_{k \to k'} \given p_{k \to k'}), & 
W_{k\to k'} &\sim \begin{cases} \distGamma(W_{k \to k'} \given \kappa, v_{k \to k'}) & \text{ if } A_{k \to k'}=1,\\
\delta_0(W_{k \to k'})  & \text{ if } A_{k \to k'} = 0.
\end{cases}
\end{align*}
The network model provides~$p_{k \to k'}$, the probability of a directed connection from node~$k$ to~$k'$, and~$v_{k \to k'}$, the inverse scale of the gamma distribution over the corresponding connection weight. We assume that~$\kappa$, the shape of the weight prior, is fixed for simplicity. The matrix~$\bA = \{\{A_{k \to k'}\}\}$ is a binary, directed adjacency matrix indicating the presence or absence of a connection for each pair of nodes, and the matrix~$\bW = \{\{W_{k \to k'}\}\}$ is a non-negative weight matrix denoting the strength of each connection.

\sloppy{The vector $G_{k \to k'}$ is a normalized probability mass function.
We model~${G_{k \to k'}[d]}$ as a convex combination of basis functions,~$\bphi_b$, which are normalized such that~${\sum_{d=1}^D \phi_b[d] \Delta t \equiv 1}$, and require~${\sum_{b=1}^B g_b^{(k,k')} \equiv 1}$ under our model. 
This constraint is implemented via a Dirichlet prior,~${\bg^{(k,k')} \sim \distDirichlet(\bgamma)}$.}

Intuitively, the weight~$W_{k\to k'}$ specifies how many child events on process~$k'$ will be caused by a single event on process~$k$. 
Then, $G_{k \to k'}[d]$ specifies the probability that the child event will occur at lag~$d \Delta t$.
In fact, this procedure is the basis of a recursive algorithm for generating samples from the discrete time Hawkes process.

Figure~\ref{fig:hawkes} illustrates the basic components of the model.
In this example, we have a stochastic block model with two types of processes (red and blue) that preferentially interact with processes of the same type.
Events induce weighted impulse responses on downstream processes according to an underlying latent network (Fig.~\ref{fig:hawkes_b}).
The impulse responses,~${G_{k\to k'}[d]}$, are modeled as a convex combination of basis functions (Fig~\ref{fig:hawkes_c}).
The impulse response is weighted according to the strength of the connection,~$W_{k \to k'}$ before being added to the rate of downstream processes.

With fixed basis functions, we can write the instantaneous discrete time in a simplified form,
\begin{align}
\label{eq:discrete_hawkes_rate_simple}
 \lambda_{t,k} &= \lambda_k^{(0)} + \sum_{k' = 1}^K  \sum_{b=1}^B W_{k' \to k} g_{b}^{(k',k)} \sum_{t'=1}^{t-1} s_{t',k'} \phi_b[t-t']\\
&= \lambda_k^{(0)} + \sum_{k' = 1}^K \sum_{b=1}^B W_{k' \to k} g_{b}^{(k',k)} \widehat{s}_{t,k',b},
\end{align}
where~${\widehat{s}_{t,k',b} \equiv (\bs_{:,k'} \ast \bphi_b)[t]}$ can be precomputed. Here, the instantaneous rate reduces to a sum of a weighted inputs, which suggests a Bayesian inference algorithm via data augmentation.

\section{Inference with Gibbs Sampling}
\begin{figure}
\centering
\includegraphics[width=2.5in]{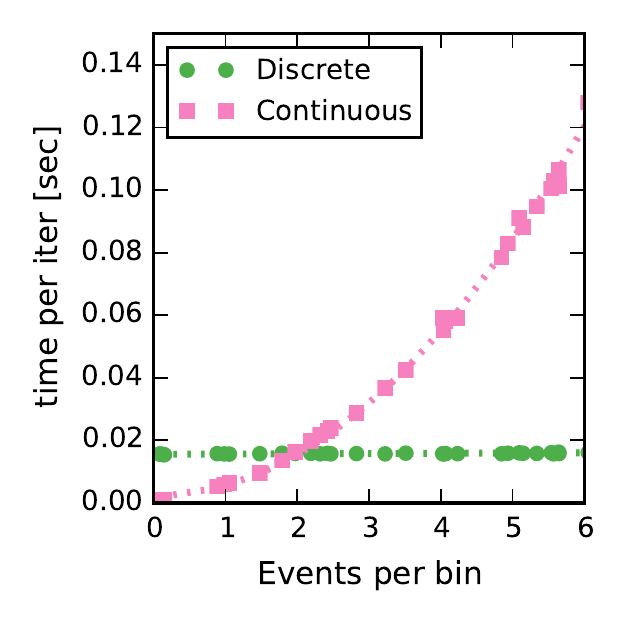}
\vskip-1em
\caption{Comparison of run time per Gibbs sweep for the discrete and continuous network Hawkes formulations. Best fit lines added.}
\label{fig:disc_vs_cont}
\end{figure}
As in other preceding work \cite{Simma-2010}, we begin by introducing auxiliary parent variables for each entry~$s_{t,k}$.
By the superposition theorem for Poisson processes, each event can be attributed to either the background rate or one of the impulse responses.

Let~${z_{t,k'}^{(k,b)} \in \{0,\ldots, s_{t,k'}\}}$ denote how many of the events that occurred in the~$t$-th time bin on the~$k'$-th process are attributed to the~$b$-th basis function of the~$k$-th process.
Similarly, let~${z_{t,k'}^{(0)}}$ denote the number of events attributed to the background process. We combine these auxiliary variables into vectors,~${\bz_{t,k'}\triangleq \left[z_{t,k'}^{(0)}, z_{t,k'}^{(1,1)}, \ldots, z_{t,k'}^{(K,B)} \right]}$.

Due to the Poisson superposition principle, these parent variables are conditionally multinomial distributed.
For time~$t$ and process~$k'$, we resample
\begin{align*}
\bz_{t,k'} &\sim \distMultinomial \left(s_{t,k'}, \bu_{t,k'} \right) &
u_{t,k'}^{(0)} &= \frac{\lambda_{k'}^{(0)}}{\lambda_{t,k'}}, &
u_{t,k'}^{(k,b)} &= \frac{W_{k \to k'} g_b^{(k,k')} \widehat{s}_{t,k,b}}{\lambda_{t,k'}}\, .
\end{align*}
Given this attribution, the likelihood factorizes into a product of Poisson distributions,
\begin{align*}
p(\bz \given \blambda) = \left[\prod_{t=1}^T \prod_{k'=1}^K \distPoisson(z_{t,k'}^{(0)} \given \lambda_{k'}^{(0)} \Delta t)\right] 
\left[ \prod_{t=1}^T \prod_{k=1}^K \prod_{k'=1}^K  \distPoisson(z_{t,k'}^{(k,b)} \given W_{k \to k'} g_b^{(k,k')} \widehat{s}_{t,k,b} \Delta t)\right].
\end{align*}

\paragraph{Gibbs sampling the background rates.}
We use conjugate priors for the constant background rates, weights, and impulse responses.
For the constant background rates we have, ${\lambda_{k'}^{(0)} \sim\distGamma(\alpha_\lambda, \beta_\lambda)}$, which results in the conditional distribution
\begin{align*}
\lambda_{k'}^{(0)} \given \{z_{t,k}^{(0)}\} &\sim
\distGamma(\alpha_\lambda^{(k)}, \beta_\lambda^{(k)}), &
\alpha_\lambda^{(k)} &= \alpha_\lambda + \sum_{t=1}^T z_{t,k'}^{(0)}, &
\beta_\lambda^{(k)} &= \beta_\lambda + T \Delta t\,.
\end{align*}

\paragraph{Gibbs sampling impulse responses.}
The likelihood of the impulse responses,~$\bg^{(k,k')}$ is proportional to a Dirichlet distribution. 
Combined with a~$\text{Dirichlet}(\bgamma)$ prior this yields
\begin{align*}
\bg^{(k,k')} &\given \{z_{t,k}^{(k',b)}\}, \bgamma  \sim \text{Dirichlet}\left( \bgamma^{(k,k')} \right), &
\gamma_b^{(k,k')} &=  \gamma_b + \sum_{t=1}^T z_{t,k'}^{(k, b)}\,.
\end{align*}

\paragraph{Gibbs sampling the weighted adjacency matrix.}
Given the adjacency matrix~$\bA$ and the parents, the weights follow a gamma distribution,
\begin{align*}
W_{k \to k'} \given A_{k \to k'}\!=\!1 &\sim \distGamma(\kappa^{(k,k')}, v^{(k,k')}), &
\kappa^{(k,k')} &= \kappa +\! \sum_{t=1}^T \sum_{b=1}^B z_{t,k'}^{(k,b)}, &
{v}^{(k,k')} &= v_{k\to k'} +\! \sum_{t=1}^T s_{t,k}.
\end{align*}
Following \citet{Linderman-2014}, to resample~$\bA$, we iterate over each entry and sample from the marginal distribution after integrating out the parents. We assume the parameters of the network prior can be sampled efficiently --- a reasonable assumption for many random network models.

The continuous time representation introduces a latent ``parent'' variable for each event in the dataset, and the parent can be any one of the events that occurred in the preceding window of influence. Call the number of potential parents~$M$. The discrete time representation has a multinomial random variable for each time bin that contains at least one event, and the support of this multinomial is always a fixed size,~${KB+1}$.  When the rate of events is high, $KB+1 \ll M$, allowing for dramatic improvements in efficiency in the discrete case. 

Figure~\ref{fig:disc_vs_cont} shows the time per full Gibbs sweep as a function of the number of events per discrete time bin for the discrete and continuous formulations. The discrete formulation incurs a constant penalty whereas the continuous formulation quickly grows with the event rate. For low rates, the continuous formulation can be advantageous, but the discrete model is vastly superior in many realistic settings. For example, \citet{Linderman-2014} worked with trades on the S\&P100, which occur tens or hundreds of times per second for each stock. Since the complexity of their algorithm grows with the number of events, they down-sampled the data to consider only the times when a stock price significantly changed.  

\section{Stochastic Variational Inference}
The discrete time formulation offers advantageous complexity compared to the continuous analogue, but it still must resample the entire set of parents each iteration in order to maintain the invariance of the posterior distribution. In many cases, a mini-batch of time bins can provide substantial information about the global parameters of the model, and rapid progress can be made by iterating quickly over subsets of the data. This motivates our derivation of a stochastic variational inference (SVI) algorithm for the network Hawkes process. 

Variational methods optimize a lower bound on the marginal likelihood by minimizing the KL-divergence between a tractable approximating distribution and the true posterior. Since the data-local variables (e.g., the parent identities) are conditionally independent given the global parameters~($\bW$,~$\bg$,~etc.), our variational approach will easily extend to the stochastic setting in which we compute unbiased estimates of the gradient of the variational objective using mini-batches of data.

The primary impediment to deriving a variational approximation is the non-conjugacy of the spike-and-slab prior on the weights. 
To overcome this, we approximate the spike-and-slab prior with a mixture of gamma distributions, as has previously explored by~\citet{Grabska-2013}:
\begin{align*}
p(A_{k \to k'}) &= \distBernoulli(A_{k \to k'} \given p_{k \to k'}),\\
p(W_{k \to k'} \given A_{k \to k'}) &=
\begin{cases} 
\distGamma(W_{k \to k'} \given \kappa, v_{k \to k'})  & \text{ if }  A_{k \to k'} =1, \\
\distGamma(W_{k \to k'} \given \kappa_0, \nu_0)  & \text{ if }  A_{k \to k'} =0.
\end{cases}
\end{align*}
As~${\kappa_0 \to 0}$ and~${\nu_0 \to \infty}$, the second mixture component converges to a delta function at zero and recovers the true spike and slab model. As we relax this prior, the weights will be nonnegative when~${A=0}$, but they will be small relative to the weights when~${A=1}$.
Importantly, with this prior the model is rendered conjugate and amenable to a matching variational factor for each pair~$(A_{k \to k'}, W_{k \to k'})$. Following \citet{Lazaro-2011}, let,
\begin{align}
q(A_{k \to k'}) &= \distBernoulli(A_{k \to k'} \given \widetilde{p}_{k \to k'}), \\
q(W_{k \to k'} \given A_{k \to k'}) &=
\begin{cases} 
\distGamma(W_{k \to k'} \given \widetilde{\kappa}_1^{(k,k')}, \widetilde{v}_1^{(k,k')})  & \text{ if }  A_{k \to k'} =1, \\
\distGamma(W_{k \to k'} \given \widetilde{\kappa}_0^{(k,k')}, \widetilde{v}_0^{(k,k')})  & \text{ if }  A_{k \to k'} =0.
\end{cases}
\end{align}
The rest of the variational approximation is fully factorized. Since the model is now conjugate, factors are easily derived. We provide a complete derivation in  Appendix~\appvariational~and state the final forms here.
\paragraph{Variational updates for parent variables,~$q(\bz_t)$} 
For the parent variables, the variational updates are
\begin{align}
q(\bz_{t,k'}) &= \distMultinomial(\bz_{t, k'} \given s_{t,k'}, \widetilde{\bu}_{t,k'}) \\
\widetilde{u}_{t,k'}^{(0)} &\propto \exp\left\{\mathbb{E}_{\blambda}[\ln \lambda_k^{(0)}]\right\}\\
\widetilde{u}_{t,k'}^{(k,b)} &\propto \widehat{s}_{t,k,b} \exp\left\{\mathbb{E}_g [\ln g_b^{(k,k')}] + \mathbb{E}_W[\ln W_{k,k'}]\right\}\,.
\end{align}

\paragraph{Variational updates for background rates,~$q(\lambda_k^{(0)})$}
The variational form parameters of the gamma distribution over background rates are
\begin{align*}
q(\lambda_k^{(0)}) &= \distGamma(\lambda_k^{(0)} \given \widetilde{\alpha}_{\lambda}^{(k)}, \widetilde{\beta}_{\lambda}^{(k)}),  &
\widetilde{\alpha}_{\lambda}^{(k)} &= \alpha_\lambda + \sum_{t=1}^T \mathbb{E}_{\bz}\left[ z_{t,k}^{(0)}\right], &
\widetilde{\beta}_{\lambda}^{(k)} &= \beta_\lambda + T \Delta t\,.
\end{align*}

\paragraph{Variational approximation for impulse response parameters,~$q(\bg^{(k,k')})$}
With the conjugate prior formulation the variational parameter updates for the Dirichlet distributed impulse response parameters are
\begin{align*}
q(\bg^{(k,k')}) &= \distDirichlet(\bg^{(k,k')} \given \widetilde{\bgamma}^{(k,k')}), &
\widetilde{\gamma}_b^{(k,k')} &= \gamma_{b} + \sum_{t=1}^T \mathbb{E}_{\bz}\left[z_{t,k'}^{(k,b)}\right]\,.
\end{align*}

\paragraph{Variational approximation for the weighted adjacency matrix.}
The primary motivation for adopting a weakly sparse mixture of gamma distributions is to derive an efficient variational inference algorithm.
The mixture-of-gammas prior is conjugate with the Poisson observations, and hence the variational distribution is also a mixture of gammas:
\begin{align*}
q(W_{k \to k'} \given A_{k \to k'}=1) &= \distGamma(W_{k \to k'} \given \widetilde{\kappa}^{(k,k')}_1, \widetilde{v}^{(k,k')}_1) \\
\widetilde{\kappa}^{(k,k')}_1 = \kappa + \sum_{t=1}^T \sum_{b=1}^B \mathbb{E}\left[z_{t,k'}^{(k,b)}\right]  \quad & \quad
\widetilde{v}^{(k,k')}_1  = \mathbb{E}[v_{k \to {k'}}] + \sum_{t=1}^T s_{t,k}\,,
\end{align*}
and likewise for the ``spike'' factor,
\begin{align*}
q(W_{k \to k'} \given A_{k \to k'}=0) &= \distGamma(W_{k \to k'} \given \widetilde{\kappa}^{(k,k')}_0, \widetilde{v}^{(k,k')}_0) \\
\widetilde{\kappa}^{(k,k')}_0 = \kappa_0 + \sum_{t=1}^T \sum_{b=1}^B \mathbb{E}\left[z_{t,k'}^{(k,b)}\right] \quad & \quad
\widetilde{v}^{(k,k')}_0  = \nu_0 + \sum_{t=1}^T s_{t,k}\,.
\end{align*}

This leaves us with~$q(A_{k \to k'})$, which is Bernoulli distributed with parameter~$\widetilde{p}_{k \to k'}$.  
Collecting all the terms that include~$A_{k \to k'}$ and lack~$W_{k \to k'}$ yields
\begin{multline*}
 \frac{\widetilde{p}_{k \to k'}}{1-\widetilde{p}_{k \to k'}} =  
 \frac{\exp\{\mathbb{E} [\ln p_{k \to k'}] \} }{\exp\{\mathbb{E}[\ln (1-p_{k \to k'})] \}} \times 
\frac{ (\exp\{\mathbb{E} [\ln v_{k \to k'}] \})^{\kappa} }{ \Gamma(\kappa)} \times 
\frac{\Gamma(\widetilde{\kappa}^{(k,k')}_1)}{ (\widetilde{v}^{(k,k')}_1)^{\widetilde{\kappa}^{(k,k')}_1} } \times
\frac{\Gamma(\kappa_0)}{ (\nu_0)^{\kappa_0} } \times
\frac{(\widetilde{v}^{(k,k')}_0)^{\widetilde{\kappa}^{(k,k')}_0}}{ \Gamma(\widetilde{\kappa}^{(k,k')}_0)}\,.
\end{multline*}

As with Gibbs sampling, we assume a variational approximation for the network model can be derived, and provide access to the necessary expectations,~$\mathbb{E}[\ln p_{k\to k'}]$, $\mathbb{E}[\ln(1-p_{k\to k'})]$, $\bbE[v_{k\to k'}]$, and~$\mathbb{E}[\ln v_{k\to k'}]$. 

As aforementioned, the time bins are conditionally independent given the network weights and the adjacency matrix --- a common pattern exploited by stochastic variational inference (SVI) algorithms \cite{Hoffman-2013}.
These methods optimize the variational objective using stochastic gradient methods that work with mini-batches of data.
Often, a mini-batch of data can provide valuable information about the global parameters, in our case the network and background rates. 
Quickly iterating over these global parameters allows us to reach good modes of the posterior distribution in a fraction of the time that batch variational Bayes and Gibbs sampling require, since those methods must process the entire dataset before making an update.
SVI does require some tuning, however. In particular, we must set a mini-batch size and a step size schedule.
In this work, we fix the mini-batch size to~${T_{\mathsf{mb}}=1024}$ and set the  step size at iteration~$i$ to~${(i+1)^{-0.5}}$.
These parameters may be  tuned with general purpose hyperparameter optimization techniques \cite{Snoek-2012}.

\section{Synthetic Results}

\begin{figure*}[t!]
\begin{center}
\begin{subfigure}[b]{0.48\linewidth}
\centering
\includegraphics[width=3in]{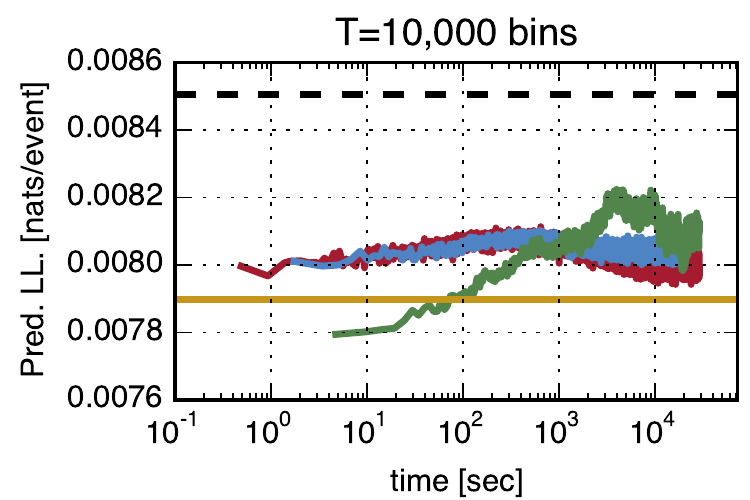} 
\end{subfigure}
~
\begin{subfigure}[b]{0.48\linewidth}
\centering
\includegraphics[width=3in]{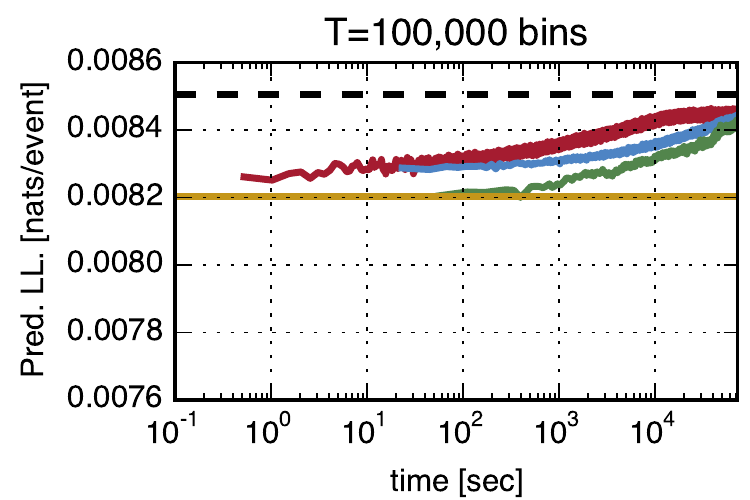} 
\end{subfigure}\\
\begin{subfigure}[b]{\linewidth}
\centering
\includegraphics[width=5.25in]{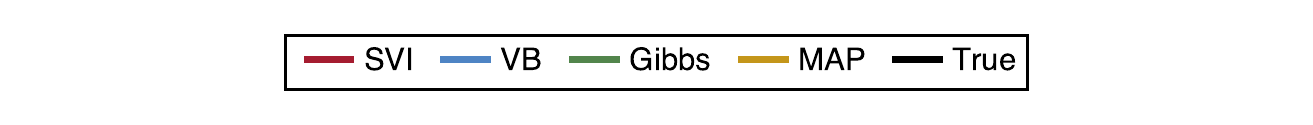} 
\end{subfigure}
\caption{Predictive log likelihood versus wall clock time for three Bayesian inference algorithms on a dataset of~$K=50$ processes and~$T=10^4$ and~$T=10^5$ time bins on the left and right, respectively.}
\label{fig:synth_pll}
\end{center}
\end{figure*}

We assess the performance of the proposed inference algorithms on a synthetic dataset generated by a strongly sparse, discrete time Hawkes process with~${K=50}$ processes. 
The network is an Erd\H{o}s-Renyi graph with uniform connection probability of~$p=0.08$, and the weights are independently and identically distributed with a~$\distGamma(3, 15)$ prior.
We simulate~${T=10^5}$ time bins in steps of size~${\Delta t=1}$.
The processes have a mean background rate of~$1.0$ event per time bin and, due to the network interactions, the average total rate of the processes is~$16.7 \pm 12.0$ events per bin.
Referring to Figure~\ref{fig:disc_vs_cont}, this is a regime that favors the discrete model.
Then we trained the strongly sparse discrete time model using Gibbs sampling, and the weakly sparse discrete time model using either batch variational Bayesian inference (VB) or stochastic variational inference (SVI) 

We evaluated the algorithms in terms of their predictive log likelihood on a held-out dataset of length~$T_{\mathsf{test}}=10^3$.
First, we trained the models on only the first~$T=10^4$ time bins of data. 
Figure~\ref{fig:synth_pll} (left) shows the predictive log likelihood as a function of wall-clock time, measured in units of nats per event improvement over a homogeneous Poisson process baseline. We initialized the Gibbs sampling and variational inference algorithms by rounding the cross-validated MAP estimate as described in the appendix.
For this relatively short dataset, SVI and batch VB converge at comparable rates, achieving competitive predictive log likelihood in a matter of minutes.
Gibbs sampling for the strongly sparse model converges at a considerably slower rate, but eventually outperforms the variational results from the weakly sparse model. 
The MAP estimate, even with cross validated regularization, underperforms the other competing algorithms.

This trend is amplified when we consider the entire training set of size~$T=10^5$.
Figure~\ref{fig:synth_pll} (right) illustrates the power of SVI in handling these large time datasets.
Considerable information about the global parameters (e.g., the network) can be gained from just a mini-batch of time points.
Hence, we can make rapid improvements in predictive log likelihood very quickly.
By contrast, each step of the Gibbs and batch VB algorithms is approximately 10 times slower, and even after computing sufficient statistics over the entire dataset, the algorithm is only able to make limited progress per iteration.

\section{Connectomics Results}
\begin{figure*}[t!]
\begin{center}
\begin{subfigure}[b]{0.32\linewidth}
\centering
\includegraphics[width=\textwidth]{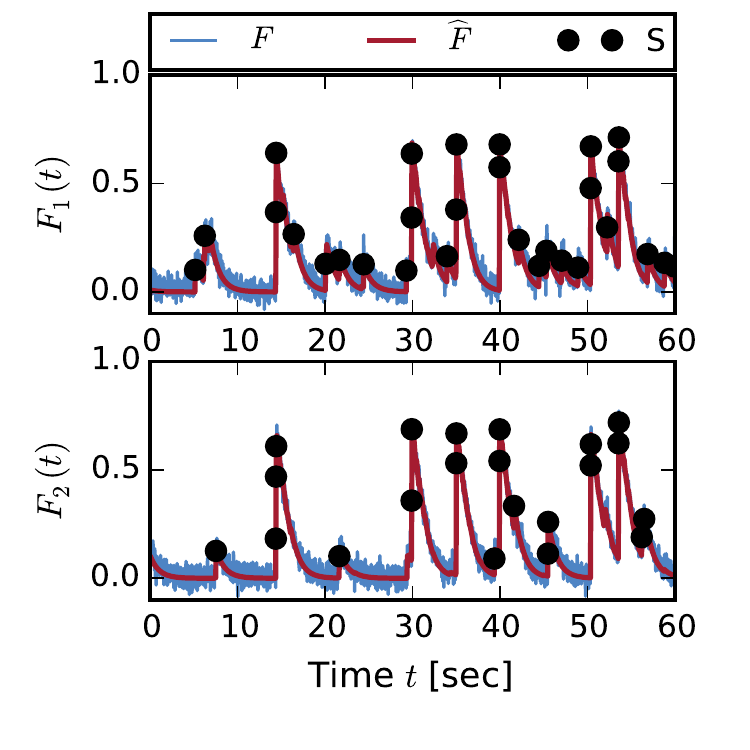} 
\vspace{-2em}
\caption{}
\label{fig:connectomics_data}
\end{subfigure}
~
\begin{subfigure}[b]{0.32\linewidth}
\centering
\includegraphics[width=\textwidth]{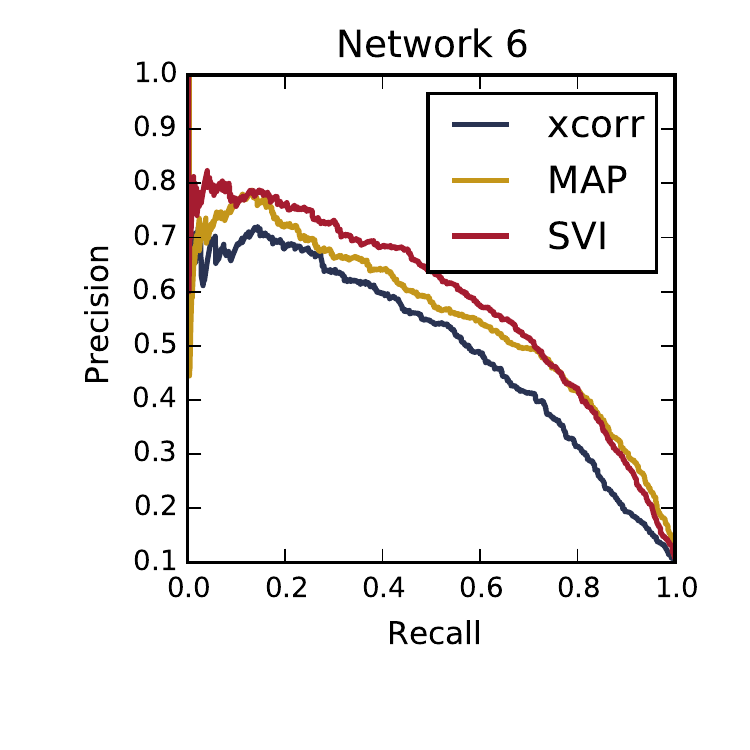} 
\vspace{-2em}
\caption{}
\label{fig:connectomics_prc}
\end{subfigure}
~
\begin{subfigure}[b]{0.32\linewidth}
\centering
\includegraphics[width=\textwidth]{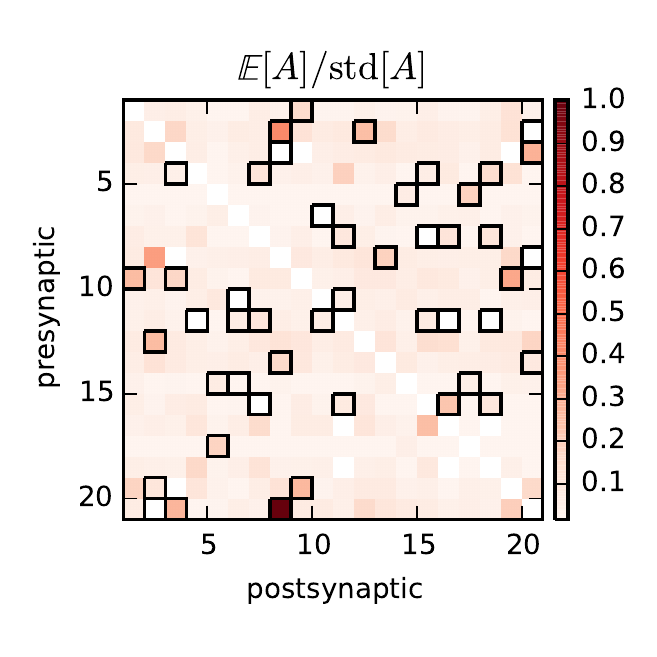} 
\vspace{-1.5em}
\caption{}
\label{fig:connectomics_zscore}
\end{subfigure}
\vspace{-1em}
\caption{Application of the network Hawkes model to a connectomics challenge. The data are in the form of a calcium fluorescence trace, which we preprocess to extract neural spike times (a). We measure performance on a link prediction task using a precision-recall curve and find that the posterior estimates of SVI provide the best estimates on some networks (b). In addition to an estimate of the connection probability and weight, SVI provides an estimate of the posterior uncertainty. (c) shows~$\mathbb{E}_q[\bA] / \text{std}_q[\bA]$ for the first 20 neurons. True connections are outlined in black.}
\label{fig:connectomics}
\end{center}
\vspace{-1em}
\end{figure*}

We tested these inference algorithms on the data from the Chalearn neural connectomics challenge\footnote{\url{http://connectomics.chalearn.org}} \cite{Stetter-2012}.
The data consist of calcium fluorescence traces,~$\bF$, from six networks of~$K=100$ neurons each. We use ten minutes of data at 50Hz sampling frequency to yield~$T=3\times 10^6$ entries in~$\bS$.
In this case, the networks are purely excitatory, and each action potential, or spike, increases the probability of the downstream neurons firing as a result.
This matches the underlying intuition of the Hawkes process model, making it a natural choice.  

In order to apply the Hawkes model, we first convert the fluorescence traces into a spike count matrix using OOPSI, a Bayesian inference algorithm based on a model of calcium fluorescence \cite{Vogelstein-2010}.
The output is a filtered fluorescence trace,~$\widehat{\bF}$, and a probability of spike for each time bin.
We threshold this at probability~$0.7$ to get a~$T\times K$ binary spike matrix,~$\bs$.
This preprocessing is shown in Figure~\ref{fig:connectomics_data}.

Figure~\ref{fig:connectomics_prc} shows the precision-recall curve we used to evaluate the algorithms' performance on network recovery. As a baseline, we compare against simple thresholding of the cross correlation matrix. On Network 6, SVI offers the best network inference. Table 5 shows the results on the other five networks using the same model parameters. On 4/5 of these networks, the Bayesian methods offer the best performance. 

Figure~\ref{fig:connectomics_zscore} illustrates one of the main advantages of the fully Bayesian inference algorithm -- calibrated estimates of posterior uncertainty. Here we show the SVI algorithm's estimate of the posterior mean of~$\bA$ normalized by the posterior standard deviation for a subset of 20 neurons from Network 6. We also outline the true connections to show that the most confident predictions are more likely to correspond to true connections. Such estimates of the posterior uncertainty are not available with standard heuristic methods or point estimates.

\begin{table*}
\centering
\resizebox{\textwidth}{!}{%
\begin{tabular}{ c||c|c||c|c||c|c||c|c||c|c }
& \multicolumn{2}{c||}{Network 1}
& \multicolumn{2}{c||}{Network 2}
& \multicolumn{2}{c||}{Network 3}
& \multicolumn{2}{c||}{Network 4}
& \multicolumn{2}{c}{Network 5} \\
\cline{2-11}
Algorithm
& ROC & PRC
& ROC & PRC
& ROC & PRC
& ROC & PRC
& ROC & PRC \\
\hhline{=#=|=#=|=#=|=#=|=#=|=}
xcorr & 0.596 & 0.139 & 0.591 & 0.133 & \bf{0.701} & \bf{0.198} & 0.745 & 0.296 & 0.798 & 0.359  \\
\hline
MAP & 0.607 & 0.174 & \bf{0.619} & \bf{0.143} & 0.698 & 0.178 & \bf{0.790} & 0.334 & \bf{0.859} & 0.408 \\
\hline
SVI  & \bf{0.649} & \bf{0.184} & 0.605 & 0.141  & 0.673 & 0.176 & 0.774 & \bf{0.342}  & 0.844 & \bf{0.410}\\
\hline
\end{tabular}
}
\caption{Comparison of inference algorithms on link prediction for five networks from the Chalearn connectomics challenge. Performance is measured by area under the ROC curve and area under the precision recall curve (PRC). In four of the five networks a Hawkes process model provides the best results.}
\end{table*}

\section{Conclusion}
We presented a scalable stochastic variational inference algorithm for the problem of Bayesian network discovery with Hawkes process observations. Building on previous modeling work, we leveraged a weak sparsity model to obtain a fully conjugate model. We focused on scaling to long duration recordings (large~$T$). Scaling to large networks (large~$K$) is nontrivial due to dependencies among weights, but in future work we hope to explore approximate algorithms to tackle this important problem.

\paragraph{Acknowledgments} We thank the members of the Harvard Intelligent Probabilistic Systems (HIPS) group, especially Matthew Johnson, for many helpful conversations. S.W.L. is supported by the Center for Brains, Minds and Machines (CBMM), funded by NSF STC award CCF-1231216.  R.P.A. is partially supported by NSF IIS-1421780.

\bibliography{arxiv}
\bibliographystyle{unsrtnat}

\appendix

 %
%
\section{Derivation of the Gibbs Sampling Algorithm}

The updates for the weights and the background rates are straightforward extensions of those presented in \cite{Linderman-2014}. The only nontrivial derivation of the Gibbs sampling is the update for the impulse response parameters. The likelihood of the impulse responses,~$\bg^{(k,k')}$ is proportional to a Dirichlet distribution. This can be seen by observing that the Hawkes process can be sampled by recursing over events. For each event we sample the number of offspring from~$\distPoisson(W_{k \to k'})$, and then we sample the offsets of those offspring from~$G_{k \to k'}[d]$. Since~$G$ is modeled as a convex combination of basis functions, the choice of basis function is essentially a draw from a categorical distribution. We can also derive this directly from the likelihood:
\begin{align*}
&p(\{\{z_{t,k'}^{(k,b)}\}_{t=1}^T\}_{b=1}^B \given \bg^{(k,k')}, \bW) \propto \prod_{t=1}^T \prod_{b=1}^B \text{Poisson}\left(z_{t,k'}^{(k, b)} \given  W_{k,k'} g_b^{(k,k')}\widehat{s}_{t,k,b} \Delta t \right) \\ 
&\propto \prod_{t=1}^T \prod_{b=1}^B \left( W_{k,k'} g_b^{(k,k')}\widehat{s}_{t,k,b} \Delta t\right)^{z_{t,k'}^{(k, b)}} \times \exp\left\{- W_{k,k'} g_b^{(k,k')}\widehat{s}_{t,k,b} \Delta t\right\} \\
& \propto  \prod_{b=1}^B \left( g_b^{(k,k')}\right)^{\sum_{t=1}^T z_{t,k'}^{k, b}} \exp\left\{- W_{k,k'} g_b^{(k,k')} \sum_{t=1}^T  \widehat{s}_{t,k,b} \right\} \\
& \propto  \prod_{b=1}^B \left( g_b^{(k,k')}\right)^{\sum_{t=1}^T z_{t,k'}^{(k, b)}} \exp\left\{-  W_{k,k'}  g_b^{(k,k')}N_k\right\} \\
& \propto  \left[ \prod_{b=1}^B \left( g_b^{(k,k')}\right)^{\sum_{t=1}^T z_{t,k'}^{(k, b)}} \right]  \exp\left\{-  W_{k,k'} N_k \sum_{b=1}^B g_b^{(k,k')}\right\} \\
& \propto  \left[ \prod_{b=1}^B \left( g_b^{(k,k')}\right)^{\sum_{t=1}^T z_{t,k'}^{(k, b)}} \right]  \exp\left\{-  W_{k,k'} N_k\right\} \\
& \propto \text{Dirichlet}\left(\bg^{(k,k')} \, \bigg| \, \left[ \sum_{t=1}^T z_{t,k'}^{(k, 1)}, \ldots, \sum_{t=1}^T z_{t,k'}^{(k, B)} \right]\right).
\end{align*}

Combined with a~$\text{Dirichlet}(\bgamma)$ prior this yields,
\begin{align*}
\bg^{(k,k')} &\given \{z_{t,k}^{(k',b)}\}, \bgamma  \sim \text{Dirichlet}\left( \bgamma' \right), \\
\gamma_b' &=  \gamma_b + \sum_{t=1}^T z_{t,k'}^{(k, b)}.
\end{align*}

%
%
\section{Derivation of the Variational Inference Algorithm}
\label{app:variational}
Our goal is to approximate the posterior distribution,~$p(\bz, \bA, \bW, \blambda^{(0)}, \bg, \thetanet \given \bs)$, with a
variational distribution,~${q(\bz, \bA, \bW, \blambda^{(0)}, \bg, \thetanet)}$,
by minimizing the KL-divergence between~$q$ and~$p$. As before, we have
introduced auxiliary parent variables,~$\bz$, to decouple the events
attributed to each potential parent process. We then restrict~$q$ to take a factorized form,
\begin{align}
q(\bz, \bA, \bW, \blambda^{(0)}, \bg, \thetanet) &= 
\prod_{t=1}^T q(\bz_t) \prod_{k=1}^K q(\lambda_k^{(0)})  
\prod_{k=1}^K \prod_{k'=1}^K q(A_{k,k'}, W_{k,k'}) q(\bg^{(k, k')}) 
q(\thetanet)
\end{align} 
This is essentially the same as a typical factorized approximation
except that we have combined~$A_{k,k'}$ and~$W_{k,k'}$ into a single
factor, as in \citet{Lazaro-2011}. This allows a multimodal
spike-and-slab approximating distribution. 

With this factorized approximation each individual factor must satisfy
a set of \emph{mean field} consistency equations in which the log of
each factor is equal (up to a constant) to the expectation of the log
posterior under the remaining variational factors. 

\paragraph{Variational approximation for parent variables,~$q(\bz_t)$} 
For the parent variables, the consistency equations imply,
\begin{align*}
\ln q(z_{t,k'}^{(0)}) &= \mathbb{E}_{\blambda^0}\left[\ln p(z_{t,k'}^{(0)}, \bA, \bW, \blambda^{(0)} \given {\bs}) \right] + \text{const.} \\
&= \mathbb{E}_{\blambda^0}\left[\ln p(z_{t,k'}^{(0)} \given \lambda_k^{(0)}) \right] + \text{const.}  \\
&= -\ln z_{t,k'}^{(0)}! - \mathbb{E}_{\blambda^0}\left[\lambda_k^{(0)}\right] + z_{t,k'}^{(0)} \mathbb{E}_{\blambda^0}\left[\ln \lambda_k^{(0)}\right] + \text{const.},
\end{align*}
and
\begin{align*}
\ln q(z_{t,k'}^{(k,b)}) &= \mathbb{E}_{\bA, \bW, \bg}\left[\ln p(z_{t,k'}^{(k,b)}, \bA, \bW, \blambda^{(0)}, \bg \given {\bs}) \right] + \text{const.} \\
&= \mathbb{E}_{\bA, \bW, \bg}\left[\ln p(z_{t,k'}^{(k,b)} \given \bA, \bW, \bg \given {\bs}) \right] + \text{const.}  \\
&= -\ln z_{t,k'}^{(k,b)}! - \mathbb{E}_{\bW,\bA,\bg}\left[W_{k,k'} g_b^{(k,k')}\right] \widehat{s}_{t,k,b}  + z_{t,k'}^{(k,b)} \mathbb{E}_{\bW, \bA, \bg}\left[\ln \left\{W_{k,k'} g_b^{(k,k')}\widehat{s}_{t,k,b} \right\} \right] + \text{const.}
\end{align*}
Combined with the constraint that~$z_{t,k'}^{(0)} + {\sum_{k,b}
  z_{t,k'}^{(k,b)} = s_{t,k'}}$, this is the form of a multinomial
distribution with variational parameter~$\widetilde{\bu}_{t,k'}$,
\begin{align}
\label{eq:qz} q(\bz_{t,k'}) &= \text{Multinomial}(\bz_{t,k'} \given s_{t,k'}, \widetilde{\bu}_{t,k'}), \\
\nonumber \widetilde{u}_{t,k'}^{(0)} &= \frac{1}{Z} \exp\left\{\mathbb{E}[\ln \lambda_k^{(0)}]\right\} \\
\nonumber \widetilde{u}_{t,k'}^{(k,b)} &= \frac{1}{Z} \widehat{s}_{t,k,b} \exp\left\{\mathbb{E}[\ln g_b^{(k,k')}]\right\} \exp\left\{\mathbb{E}[\ln W_{k,k'}]\right\} \\
\nonumber Z &= \exp\left\{\mathbb{E}[\ln \lambda_k^{(0)}]\right\} +  \sum_{k=1}^K \sum_{b=1}^B \widehat{s}_{t,k,b} \exp\left\{\mathbb{E}[\ln g_b^{(k,k')}]\right\} \exp\left\{\mathbb{E}[\ln W_{k,k'}]\right\}.
\end{align}

\paragraph{Variational approximation for impulse response parameters,~$q(\bg^{(k,k')})$}
The consistency equations yield,
\begin{align*}
\ln q(\bg^{(k,k')}) &= \mathbb{E}_{\bz, \bW}\left[\ln p(\bz_{t,k}, \bW, \bg \given {\bs}) \right] + \text{const.} \\
&=  \sum_{b=1}^B  \left(\gamma_b + \sum_{t=1}^T \mathbb{E}_{\bz}\left[z_{t,k'}^{(k,b)}\right] \right) \ln g_b^{(k,k')}+ \text{const.}
\end{align*}
With the conjugate prior formulation the variational approximation is
again a Dirichlet,
\begin{align}
\label{eq:qbeta}
q(\bg^{(k,k')}) &= \text{Dirichlet}(\widetilde{\bgamma}^{(k,k')}) &
\widetilde{\gamma}_b^{(k,k')} &= \gamma_{b} + \sum_{t=1}^T \mathbb{E}_{\bz}\left[z_{t,k'}^{(k,b)}\right].
\end{align}

\paragraph{Variational approximation for constant background rates,~$q(\lambda_k^{(0)})$}
The variational form for~$q(\lambda_k^{(0)})$ is also determined by the
conjugate model. We have,
\begin{align*}
\ln q(\lambda_k^{(0)}) &= \mathbb{E}_{\bz}\left[\ln p(\bz_{t,k}, \bA, \bW, \blambda^{(0)}, \bg \given {\bs}) \right] + \text{const.} \\
&=  \sum_{t=1}^T -\lambda_k^{(0)} \Delta t + \mathbb{E}_{\bz}\left[z_{t,k}^{(0)}\right] \ln \lambda_k^{(0)} + (\alpha_\lambda -1)\ln \lambda_k^{(0)}  - \beta_\lambda \lambda_k^{(0)} + \text{const.}
\end{align*}
This is the form of a gamma distribution with variational parameters,
\begin{align}
\label{eq:qlambda}
q(\lambda_k^{(0)}) &= \distGamma(\widetilde{\alpha}_\lambda^{(k)}, \widetilde{\beta}_\lambda^{(k)}) &
\widetilde{\alpha}_\lambda^{(k)} &= \alpha_\lambda + \sum_{t=1}^T \mathbb{E}_{\bz}\left[ z_{t,k}^{(0)}\right] &
\widetilde{\beta}_\lambda^{(k)} &= \beta_\lambda + T \Delta t.
\end{align}

\paragraph{Variational approximation for spike-and-slab weights,~$q(A_{k,k'},W_{k,k'})$}
Returning to the variational factor for~$\bz$ in Equation~\ref{eq:qz},
we see that a problem arises with the spike-and-slab model. If our
model and our variational approximation are to share a spike-and-slab
formulation then the expected log weight will be,
\begin{align*}
\mathbb{E}_{\bW}[\ln W_{k,k'}] &= \mathbb{E}_{\bA} \mathbb{E}_{\bW \given \bA}[\ln W_{k,k'}] 
= p \mathbb{E}[\ln W_{k,k'} \given A_{k,k'}=1] + (1-p) \ln 0 = -\infty.
\end{align*}
To avoid this degeneracy, we replace the delta function with a gamma
distribution,~$p(W_{k,k'}\given A_{k,k'}=0)=\distGamma(\kappa_0,
v_0)$, which converges to a delta function as~$\kappa_0 \to 0$
and~$v_0 \to \infty$.  The prior on weights is then a mixture of
two gamma distributions, and is conjugate with the Poisson
observations. This in turn implies a variational distribution that is
also a mixture of gammas. We derive this with the consistency
equations,
\begin{align*}
\ln q(&A_{k,k'}, W_{k,k'}) \\
&= \mathbb{E}\left[\ln p(\bz_{t,k'}, \bA, \bW, \bg, \thetanet \given {\bs}) \right] + \text{const.} \\
&=  \sum_{t=1}^T \sum_{b=1}^B -W_{k,k'} \mathbb{E}_{\bg}\left[ g_b^{(k,k')}\right] \widehat{s}_{t,k,b} \, \Delta t+ \mathbb{E}_{\bz}\left[z_{t,k'}^{(k,b)}\right] \ln  W_{k,k'} \\
& \quad + A_{k,k'}\bigg[ \kappa \, \mathbb{E}_{\thetanet}[\ln v_{k \to {k'}}] - \Gamma(\kappa) + (\kappa-1) \ln W_{k,k'} - \mathbb{E}_{\thetanet} [v_{k \to {k'}}] W_{k,k'} + \bbE_{\thetanet}[\ln p_{k\to k'}] \bigg] \\
& \quad + (1-A_{k,k'}) \bigg[ \kappa_0 \ln v_0 - \Gamma(\kappa_0)  (\kappa_0-1) \ln W_{k,k'} - v_0 W_{k,k'} + \bbE_{\thetanet}[\ln(1-p_{k\to k'})] \bigg] + \text{const.}
\end{align*}
As expected,~$q(W_{k,k'} \given A_{k,k'})$ has the form of a gamma
distribution,
\begin{align}
\label{eq:qwa}
q(W_{k,k'} &\given A_{k,k'}=1) = \distGamma(W_{k,k'} \given \kappa_1^{(k,k')}, v_1^{(k,k')}) \\
\nonumber \kappa_1^{(k,k')} &= \kappa + \sum_{t=1}^T \sum_{b=1}^B \mathbb{E}_{\bz}\left[z_{t,k'}^{(k,b)}\right]  \\
\nonumber v_1^{(k,k')} &= \mathbb{E}_{\thetanet} [v_{k \to {k'}}] +  \sum_{t=1}^T \sum_{b=1}^B \mathbb{E}_{\bg}\left[g_b^{(k,k')}\right]  \widehat{s}_{t,k,b} \Delta t \\
\nonumber  &= \mathbb{E}_{\thetanet} [v_{k \to {k'}}] + N_k \sum_{b=1}^B \mathbb{E}_{\bg}\left[g_b^{(k,k')}\right],
\end{align}
and
\begin{align}
q(W_{k,k'} &\given A_{k,k'}=0) = \distGamma(W_{k,k'} \given \kappa_0^{(k,k')}, v_0^{(k,k')}) \\
\nonumber \kappa_0^{(k,k')} &= \kappa_0 + \sum_{t=1}^T \sum_{b=1}^B \mathbb{E}_{\bz}\left[z_{t,k'}^{(k,b)}\right] \\
\nonumber v_0^{(k,k')} &= v_0 +  \sum_{t=1}^T \sum_{b=1}^B \mathbb{E}_{\bg}\left[g_b^{(k,k')}\right] \widehat{s}_{t,k,b} \Delta t\\
\nonumber &= v_0 +  N_k \sum_{b=1}^B \mathbb{E}_{\bg}\left[g_b^{(k,k')}\right].
\end{align}

This leaves us with~$q(A_{k,k'})$, which we take to be Bernoulli
distributed with parameter~$\widetilde{p}_{k,k'}$. This implies,
\begin{multline*}
A_{k,k'}\left[\ln \widetilde{p}_{k \to k'} + \ln \distGamma(\widetilde{\kappa}_1^{(k,k')}, \widetilde{v}_1^{(k,k')})\right] + (1-A_{k,k'})\left[\ln (1-\widetilde{p}_{k \to k'}) + \ln \distGamma(\widetilde{\kappa}_0^{(k,k')}, \widetilde{v}_0^{(k,k')})\right] = \\
A_{k,k'}\left[\bbE_{\thetanet} [\ln p_{k \to {k'}}] + \bbE_{\thetanet} [\ln \distGamma(\kappa, v_{k \to {k'}})]\right] +
(1-A_{k,k'})\left[\bbE_{\thetanet} [\ln (1 - p_{k \to {k'}})]  + \ln \distGamma(\kappa_0, v_0)\right].
\end{multline*}
Collecting all the terms that include~$A_{k,k'}$ and lack~$W_{k,k'}$ yields,
\begin{multline}
 \frac{\widetilde{p}_{k \to k'}}{1-\widetilde{p}_{k \to k'}} =  \\
 \frac{\exp\{\bbE_{\thetanet} [\ln p_{k \to {k'}}] \} }{\exp\{\bbE_{\thetanet}[\ln (1-p_{k \to {k'}})] \}} \times 
\frac{ (\exp\{\bbE_{\thetanet}[\ln v_{k \to {k'}}] \})^{\kappa} }{ \Gamma(\kappa)}  
\times \frac{\Gamma(\widetilde{\kappa}_1^{(k,k')})}{ (\widetilde{v}_1^{(k,k')})^{\widetilde{\kappa}_1^{(k,k')}}} \times 
\frac{\Gamma(\kappa_0)}{ (v_0)^{\kappa_0} } \times
\frac{(\widetilde{v}_0^{(k,k')})^{\widetilde{\kappa}_0^{(k,k')}}}{ \Gamma(\widetilde{\kappa}_0^{(k,k')})}.
\end{multline}

\paragraph{Variational updates for the network model}
In the above derivations, the only requirement on the network model is that it provide the following expectations:~$\bbE[p_{k \to k'}]$, $\bbE[\ln p_{k \to k'}]$, $\bbE[\ln(1-p_{k \to k'})]$, ~$\bbE[v_{k \to k'}]$, and~$\bbE[\ln v_{k \to k'}]$. It may make sense to fix one of these values; for example, we may have a good empirical estimate of the weight scale,~$v_{k \to k'}$, and therefore do not need to model its posterior distribution. Alternatively, we may know the overall sparsity but be uncertain of the scale. For some models, computing the necessary variational expectations and deriving the variational updates is straightforward. For example, a stochastic block model \cite{Nowicki-2001} with a gamma prior on the scale is fully conjugate and admits closed form updates.

\paragraph{Completing the variational expectations}
Now we can compute the necessary expectations for the variational
parameter updates. These are all available in closed form,
\begin{align*}
\mathbb{E}\left[ \ln \lambda_k^{(0)} \right] &= \psi(\widetilde{\alpha}_{\lambda}^{(k)} ) - \ln \widetilde{\beta}_{\lambda}^{(k)} , \\
\mathbb{E}\left[\ln W_{k,k'}\right] &= \widetilde{p}_{k \to k'}\left(\psi(\widetilde{\kappa}^{(k,k')}_1) - \ln \widetilde{v}^{(k,k')}_1 \right) + 
(1-\widetilde{p}_{k \to k'})\left(\psi(\widetilde{\kappa}^{(k,k')}_0) - \ln \widetilde{v}^{(k,k')}_0\right), \\
\mathbb{E} \left[\ln g_b^{(k,k')}\right] &= \psi\left(\widetilde{\gamma}_b^{(k,k')}\right) - \psi \left(\sum_{b'=1}^B  \widetilde{\gamma}_{b'}^{(k,k')} \right),
\end{align*}
where~$\psi(\cdot)$ denotes the digamma function. To complete
the variational updates for the background rate, impulse response, and weight parameters we have,
\begin{align*}
\mathbb{E}\left[z_{t,k'}^{(0)}\right] &= \widetilde{u}_{t,k'}^{(0)} s_{t,k'}, &
\mathbb{E}\left[z_{t,k'}^{(k,b)}\right] &= \widetilde{u}_{t,k'}^{(k,b)} s_{t,k'},
\end{align*}
and
\begin{align*}
\mathbb{E}\left[g_b^{(k,k')}\right] &= \frac{\widetilde{\gamma}_b^{(k,k')}}{\sum_{b'=1}^B \widetilde{\gamma}_{b'}^{(k,k')}}.
\end{align*}
With this final expectation, the parameter updates for the variational
weight distributions simplify
to~${\widetilde{v}^{(k,k')}_0=v_0 + N_k}$ and~${\widetilde{v}^{(k,k')}_1=\mathbb{E} [v_{k \to {k'}}] + N_k}$.

\subsection{Initialization}
\label{sec:initialization}
Variational inference algorithms can be quite sensitive to initialization of their model parameters. 
A poor initialization may converge to a suboptimal mode of the posterior distribution.
In order to initialize our algorithm, we first fit a standard (log concave) Hawkes process using MAP estimation on a subset of the data.
We use an exponential prior on the weights, equivalent to L1 regularization, and we tune the scale  of the prior with cross validation.
To covert the weights inferred under the standard Hawkes model to a sample of the network Hawkes model, we keep the largest~$p$-fraction of the weights and set the remainder to zero.
Finally, we initialize the variational parameters of the network Hawkes model such that they are peaked at the sample values.
For example, we set~${\widetilde{v}_1^{(k,k')}=100}$ and~${\widetilde{\kappa}_1^{(k,k')}=100 \cdot W_{k \to k'}}$.

\end{document}